\documentclass{elsart}

\usepackage{graphicx}

\usepackage{amssymb}

\usepackage{url}

\newcommand{\MI}[1]{M\!I\left(#1\right)}

\begin{document}
\sloppy
\begin{frontmatter}


\title{Resampling methods for parameter-free and robust feature selection with mutual information}

\author[CESAME]{{D. Fran\c{c}ois}}, 
\author[INRIA] {F. Rossi},
\author[CESAME]{V. Wertz} and
\author[DICE]  {M. Verleysen\corauthref{correspondingauthor}}

\corauth[correspondingauthor]{Corresponding author\\	
Email addresses: \texttt{francois@inma.ucl.ac.be} (D. Fran\c{c}ois), \texttt{Fabrice.Rossi@inria.fr} (F. Rossi), \texttt{wertz@inma.ucl.ac.be} (V. Wertz), \texttt{verleysen@dice.ucl.ac.be} (M. Verleysen)\\
URLs: \texttt{http://www.ucl.ac.be/mlg} (D. Fran\c{c}ois, V.Wertz, M. Verleysen)
\texttt{http://apiacoa.org/index.en.html} (F. Rossi)}

\address[CESAME]{Universit\'e catholique de Louvain, Machine Learning Group, CESAME \\Av. Georges Lemaitre, 4, B-1348 Louvain-la-Neuve, Belgium}
\address[INRIA]{Projet AxIS, INRIA\\ Domaine de Voluceau, Rocquencourt, B.P. 105, 78153 Le
Chesnay Cedex, France}
\address[DICE]{Universit\'e catholique de Louvain, Machine Learning Group, DICE\\Place du Levant, 3, B-1348 Louvain-la-Neuve, Belgium}

\begin{abstract}
  Combining the mutual information criterion with a forward feature selection strategy
  offers a good trade-off between optimality of the selected feature subset
  and computation time. However, it requires to set the parameter(s) of the
  mutual information estimator and to determine when to halt the forward procedure.
  These two choices are difficult to make because, as the dimensionality of the
  subset increases, the estimation of the mutual information becomes less and less reliable. This paper
  proposes to use resampling methods, a K-fold cross-validation and the
  permutation test, to address both issues. The resampling methods bring
  information about the variance of the estimator, information which can then
  be used to automatically set the parameter and to calculate a threshold to
  stop the forward procedure. The procedure is illustrated on a synthetic
  dataset as well as on real-world examples.
\end{abstract}

\end{frontmatter}

\renewcommand{\labelitemi}{-}

\section{Introduction}\label{sec:introduction}

Feature selection consists in choosing, among a set of input features, or
variables, the subset of features that has maximum prediction power for the
output. More formally, let us consider ${\bf X} = (X_1, \cdots, X_d)$ a random input
 vector and $Y$ a continuous random output variable that has to be predicted from
${\bf X}$. The task of feature selection consists in finding the features
$X_i$ that are most relevant to predict the value of $Y$ \cite{Guyon:2003}.

Selecting features is important in practice, especially when distance-based methods like k-nearest neighbors
(k-NN), Radial Basis Function Networks (RBFN) and Support Vector Machines
(SVM) (depending on the kernel) are considered. These methods are indeed   
quite sensitive to irrelevant inputs: their performances tend to decrease when useless variables
are added to the data.

When the data are high-dimensional (i.e. the initial number of variables is large) 
the exhaustive search of an optimal feature set is of course intractable. In such cases, furthermore, most methods that `work backwards' by eliminating useless features perform badly. 
The backward elimination procedure for instance, or pruning methods for the MultiLayer Perceptron \cite{Verikas:2002}, SVM-based feature selection \cite{Fung:2004}, or weighting methods like the Generalized Relevance Learning Vector Quantization algorithm \cite{Hammer:2002} require building a model with all initial features. With high-dimensional data,  this will often lead to large computation times, overfitting, convergence problems, and, more generally, issues related to the curse of dimensionality. These approaches are furthermore bound to a specific prediction model.

By contrast, a forward feature selection procedure can be applied using any model and begins with small feature subsets. Such procedure is furthermore simple and often efficient.
Nevertheless, when data are high-dimensional, it becomes difficult to perform the forward search using the prediction model directly. This is because, for every candidate feature subset, a prediction model must be fit, involving resampling techniques and grid searching for optimal structural parameters. A cheaper alternative is to estimate the relevance of each candidate subset with a statistical or information-theoretic measure, without using the prediction model itself.

The combined use of a forward feature search and an information-theoretic-based relevance criterion 
is generally considered to be a good option, when nonlinear effects prevent from using the correlation coefficient \cite{guyon:2006}. In this context, the mutual information estimated using a nearest neighbour-based approach has been shown to be effective \cite{VanDijk:2006,CILS2006}.
Nevertheless, this approach, just like most feature selection
methodologies, faces two difficulties.

The first one, which is generic for all
feature selection methods, lies in the optimal choice of the number of
features to select.  Most of the time, the number of features to select is
chosen a priori or so as to maximize the relevance criterion. The former
approach leaves no room for optimization, while the latter may be very sensitive 
to the estimation of the relevance criterion. 

The second difficulty concerns the
choice of parameter(s) in the estimation of the relevance criterion. Indeed, most of these criteria,
except maybe for the correlation coefficient, have at least one structural
parameter, like a number of units or a kernel width in a prediction model, a
number of neighbours or a number of bins in a nonparametric relevance
estimator, etc. Often, the result of the selection highly depends on the value
of that (those) parameter(s).

The aim of this paper is to provide an automatic procedure to choose the two
above-mentioned important parameters, i.e. the number of features to select in
the forward search and the structural parameter(s) in the relevance criterion
estimation.  This procedure will be detailed in a situation where the mutual
information is used as relevance criterion, and is estimated through nearest
neighbours.  Resampling methods will be used to obtain this automatic
choice. Those methods increase the computational cost of the forward search, but provide meaningful information about the quality of the estimations and the setting of parameters: it will be shown that a permutation test can be used to automatically stop the forward procedure, and that a combination of permutation and K-fold resampling allows choosing the optimal number of neighbors in the estimation of the mutual information.

The remaining of this paper is organized as follows. Section \ref{sec:priorart} introduces the mutual information, the permutation
test and the K-fold resampling, and briefly reviews how they can be used
together. Section \ref{sec:problems} illustrates the challenges in choosing the number of neighbours in the mutual information estimation and the number of features to select in a forward search. Section \ref{sec:proposedapproach} then presents the proposed
approach. The performances of the
method on real-world data are reported in Section \ref{sec:experiments}.

\section{Prior art}\label{sec:priorart}

\subsection{Mutual information-based forward feature selection}

The mutual information is a nonparametric, nonlinear, measure of relevance derived
from information theory.  Unlike correlation that only considers linear
relationships between variables, the mutual information is
theoretically able to identify relations of any type. It furthermore makes no
assumption about the distribution of the data.

The mutual information of two random variables $Z_1$ and $Z_2$ is a measure of how $Z_1$ depends on $Z_2$ and \emph{vice versa}. It can be defined from the entropy $H(.)$:
\begin{equation}\label{eq:mientropy}
\MI{Z_1;Z_2} = H(Z_1) + H(Z_2) - H(Z_1,Z_2) = H(Z_1) - H(Z_2|Z_1),
\end{equation} 
where $H(Z_2|Z_1)$ is the \emph{conditional} entropy of $Z_2$ given $Z_1$.  In that sense, it measures the loss of entropy (i.e. reduction of uncertainty) of $Z_2$ when $Z_1$ is known. If $Z_1$ and $Z_2$ are independent, $H(Z_1,Z_2) = H(Z_1) + H(Z_2)$, and $H(Z_2|Z_1) = H(Z_2)$. In consequence, the mutual information of two independent variables is zero.

For a continuous random variable $Z_1$, the entropy is defined as 
\begin{equation}
H(Z_1) = - \int p_{Z_1}(\zeta_1) \, \log\,p_{Z_1}(\zeta_1) \, {\rm d}\zeta_1,
\end{equation}
where $p_{Z_1}$ is the probability distribution of $Z_1$. Consequently, the mutual information can be rewritten, for continuous $Z_1$ and $Z_2$, as 
\begin{equation}\label{eq:midivergency}
\MI{Z_1;Z_2} = \int\!\!\!\int p_{Z_1,Z_2}(\zeta_1, \zeta_2) \, \log \, \frac{p_{Z_1,Z_2}(\zeta_1, \zeta_2)}{p_{Z_1}(\zeta_1)\cdot p_{Z_2}(\zeta_2)}\,{\rm d}\zeta_1{\rm d}\zeta_2.
\end{equation} 
It corresponds to the Kullback-Leibler distance between $p_{Z_1,Z_2}(\zeta_1, \zeta_2)$, the joint probability density of $Z_1$ and $Z_2$, and the product of their respective marginal distributions. In the discrete case, the integral is replaced by a finite sum.

In practice, the mutual information has to be estimated from the dataset, as
the exact probability density functions in the above equations are not
known. The most sensitive part of the estimation of the mutual information is
the estimation of the joint probability density function $p_{Z_1,Z_2}(\zeta_1,
\zeta_2)$. Several methods have been developed in the literature to estimate
such joint densities: histograms, kernel-based methods and splines
\cite{Scott:1992}. All those estimators depend on at least one parameter that has to be
chosen appropriately. 

In the context of a forward procedure, the mutual information is estimated between a \textbf{set} of inputs $X_i$ (instead of a single variable $X_i$) and the output $Y$.  The above definitions of entropy and mutual information remain valid, provided that $Z_1$ is replaced by a multi-dimensional variable.  The dimension of the latter grows at each iteration of the forward procedure.  Therefore the estimations of the $p_{Z_1}$ and $p_{Z_1,Z_2}$ densities must also be performed in spaces of increasing dimension.

Unfortunately, most of the density estimation methods require a sample whose size grows
exponentially with both the dimension of $Z_1$ and the dimension of $Z_2$ to provide an accurate
estimation. This is sometimes referred to as one instance of the curse of
dimensionality \cite{Bellman:1961}. In practice, one seldom has the required
number of points for an accurate estimation when the dimension is above 10.
For dimensions below or close to that value, the estimation of the
multi-dimensional mutual information can be performed with classical
multivariate density estimators \cite{Bonnlander:1994,Kwak:2002b}. With more
than 10 dimensions the estimation becomes quite unreliable with
those estimators. However, nearest neighbor-based density estimators have
been reported to be less sensitive to dimensionality than many others
\cite{Kraskov:2004,Rossi:2005} and are therefore more suitable for the forward
search strategy.

The forward search is incremental and ``greedy'' in the sense that the method
makes final decisions about features at each iteration: once a feature is
chosen, its relevance is never questioned again.  The forward search will therefore
perform at most $O(d^2)$ estimations of the criterion (rather than $2^d$ for the exhaustive search). The forward search begins with an empty set of features and adds at each iteration
the feature that has the most positive influence on the criterion. The procedure
is halted either when the \emph{a priori} chosen number of features has been selected or when adding one more feature does not improve the relevance criterion. 
 
Combining a forward search procedure with a mutual information estimator for
the relevance criterion is an idea dating back to 1994 \cite{Battiti:1994}. Before the
nearest neighbor estimator was popularized by Kraskov et al. \cite{Kraskov:2004}, the multivariate mutual information measures were most often approximated using combinations of bi-variate 
\cite{Battiti:1994,Kwak:2002a} or tri-variate \cite{Fleuret:2004} mutual
information estimations. Those approximations, however, do not estimate the
true value of the mutual information between the set of $X_i$ and $Y$, and make strong independence assumptions
between the input features. The forward strategy with the mutual information
estimated using nearest neighbors was shown to be successful \cite{Rossi:2005}
and is used as the foundation method in the present paper. It however requires manual tuning of the number of neighbors and comparisons between the respective mutual informations between sets of features of different sizes and the output, which is not always advisable in practice, as detailed in Section \ref{sec:problems}.

\subsection{Resampling methods}
Additional information is needed to select a priori sound values (i) for the structural
parameter of the estimator and (ii) for the number of selected features in the subset,
without optimizing these numbers with respect to the prediction performances
of the model. This additional information, namely an estimation of the
variance of the estimator, is brought by two resampling methods: the
permutation resampling and the K-fold resampling. 

Resampling methods have heavy computational requirements that increase the
time needed to perform the forward selection procedure. However, the running
time of the scheme proposed in Section \ref{sec:proposedapproach} remains
acceptable compared to the computational burden of alternate solutions that could
be used to choose the number of features and the parameter of the estimator
(e.g. optimizing those elements based on the performances of a prediction
model). 

It should be noted that bootstrap resampling, while generally advisable for
exploring the behavior of an estimator, is not adapted to the $k$ nearest
neighbors estimator \cite{Kraskov:2004} used in this paper. When a bootstrap
sample is generated from the original dataset, it contains duplicates of many
of the observations. As a consequence, the $k$ nearest neighbors of each
observation may contain this observation itself (sometimes even
repeated), which leads to a strong overestimation of the mutual information.

\subsubsection{K-fold resampling}

The K-fold resampling is very similar to the K-fold cross-validation scheme used for validating prediction models, except that it is used in an unsupervised manner. Given $z_1$ and $z_2$ respectively realizations of $Z_1$ and $Z_2$, and some statistic $\theta$, it consists in computing the $K$ estimates $\hat\theta_k$ of $\theta$ where one (or several) data element(s) has(ve) been removed from the analysis. Typically, the sample is partitioned into $K$ clusters of roughly equal size, and the statistic is estimated $K$ times on the sample from which the $K$th cluster was excluded. The average of those estimations is often found to be a more robust estimator of  $\theta$, while the variance of the estimations gives an idea of the sensitivity of the estimator to the particular sample.

\subsubsection{The permutation test or randomized resampling}

The permutation test \cite{good} is a nonparametric hypothesis test over some estimated statistic $\hat\theta$ involving $z_1$ and $z_2$. The statistic $\hat\theta$ can be a difference of means in a classification context, or a correlation, or, as in this paper, a mutual information. Let $\hat\theta$ be the estimation of the statistics for the given $z_1$ and $z_2$, both vectors of size $n$ drawn from $p_{Z_1}$ and $p_{Z_2}$ respectively. The aim of the test is to answer the following question : how likely is the value $\hat\theta$ given the vectors $z_1$ and $z_2$ if we suppose that $Z_1$ and $Z_2$ are independent? In particular, the value of the mutual information under such hypothesis should be zero.

The permutation test considers the empirical distribution of $z_1$ and $z_2$ to be fixed, as well as the sample size.  The random variable of interest is the value of the statistic $\theta$. In such a framework, the distribution of $\hat\theta$ is the set of all values of $\hat\theta_k$ for all $n!$ possible permutations of the elements of the vector $z_1$, or, equivalently, all permutations of the elements of the vector $z_2$. The P-value $\alpha$ associated to the test is the proportion of $\hat\theta_k$ that are larger than the value of $\hat\theta$ estimated with $z_1$ and $z_2$ without permutation.

In practice, it is not necessary to perform all $n!$ permutations. Several tens or hundreds of them are randomly performed. In this case, the exact P-value cannot be known but  a 95\% confidence interval around the observed P-value can be estimated \cite{Opdyke:2003}.

\subsection{Combined uses}

The permutation test has been extensively used in conjunction with the mutual information to perform a nonparametric statistical test of independence of variables or signals. It has been of much use in identifying nonlinear relationships between pairs of variables in exploratory analysis \cite{Craddock:2006,Hahn:2005,Hummel:2005,Purushothaman:2005,Hoffman:2003}, and to test serial independence in time series \cite{Diks:2005}.

The permutation test has also been used specifically to filter out features,
by measuring independence via mean differences, student statistics, or 
chi-squared measures. The test is used, for instance, to discard features
for which the independence hypothesis cannot be statistically rejected
\cite{Conrad:2004}, or to rank features according to the p-value estimated by
the permutation test \cite{Radivojac:2004}.  The permutation test can also be
used in the process of building a decision tree, to choose the features that
should be used at a split-point \cite{Frank:1998}.

Feature filtering with the mutual information and the permutation test was also recently proposed  \cite{VanDijk:2006,Francois:2006,Radivojac:2004}, in a pure feature ranking approach where the permutation test is used to automatically set a threshold on the value of the mutual information. 

Resampling approaches similar to the K-fold resampling (Jackknife, bootstrap, etc.) have also been used to get better estimates of the mutual information \cite{Zhou:2004} and to choose among several estimators (nearest neighbor-based, histogram-based, spline-based, etc) to estimate the mutual information between EEG signals \cite{Nicolaou:2005}. The estimator that is chosen is the one that is most robust with respect to resampling, i.e. that has the lowest variance around the estimated value.

Mutual information with permutation testing has thus been used for automatic
feature filtering, that is for discarding features that are statistically
non-relevant for the prediction. This approach however  selects many features,
more than necessary since redundancy in the features is not considered. That
is why automatic forward selection is preferable to actually select features
rather than discarding them. Furthermore, in choosing the value of the estimator structural 
parameter and the number of variables to consider in the forward search, we should not only consider the variance of the estimator but also,
and more importantly, how well it discriminates dependent features (with $M\!I>0$) from independent ones (with $M\!I=0$). The methodology described in the next section answers these questions.

\section{The sensitivity to parameter values}\label{sec:problems}
The mutual information, with a nearest neighbor-based estimator, and the
forward search combined together present a good compromise between computation
time and performances. As already discussed, two issues must be addressed however, namely the number
of features to select and the choice of the parameter in the estimation of the
mutual information to discriminate at best relevant features from useless ones. 
The results of the feature selection process highly depend on those two parameters,
especially when the mutual information must be estimated from a few samples.
This section illustrates those difficulties in a simple case.

The problem discussed here is a synthetic prediction problem, derived from
Friedman's \cite{friedman}. We consider 10 input variables $X_i$ and one
output variable $Y$ given by
\begin{equation}\label{eq:friedman}
Y = 10 \sin\left(X_1 \cdot X_2\right) + 20 \left( X_3-0.5\right)^2 + 10 X_4 + 5 X_5 + \epsilon.
\end{equation}
All $X_i, 1\leq i\leq 10$, are uniformly distributed over $[0,1]$, and
$\epsilon$ is a centered Gaussian noise with unit variance. Variables $X_6$ to
$X_{10}$ are just noise and have no predictive power. The sample size is 100.

\subsection{Parameter sensitivity}

The number $k$ of neighbors taken into account in the estimation of the mutual
information must be chosen carefully, especially in the case of a small sample
and noisy data. If the number of neighbors is too small, the estimation will
have a large variance; if the number of neighbors is chosen too large,
the variance of the estimator will be  small, but all estimations will
converge to zero, even for highly-dependent variables.

\begin{figure}[htbp]
\centering
\includegraphics[width=1.0\textwidth]{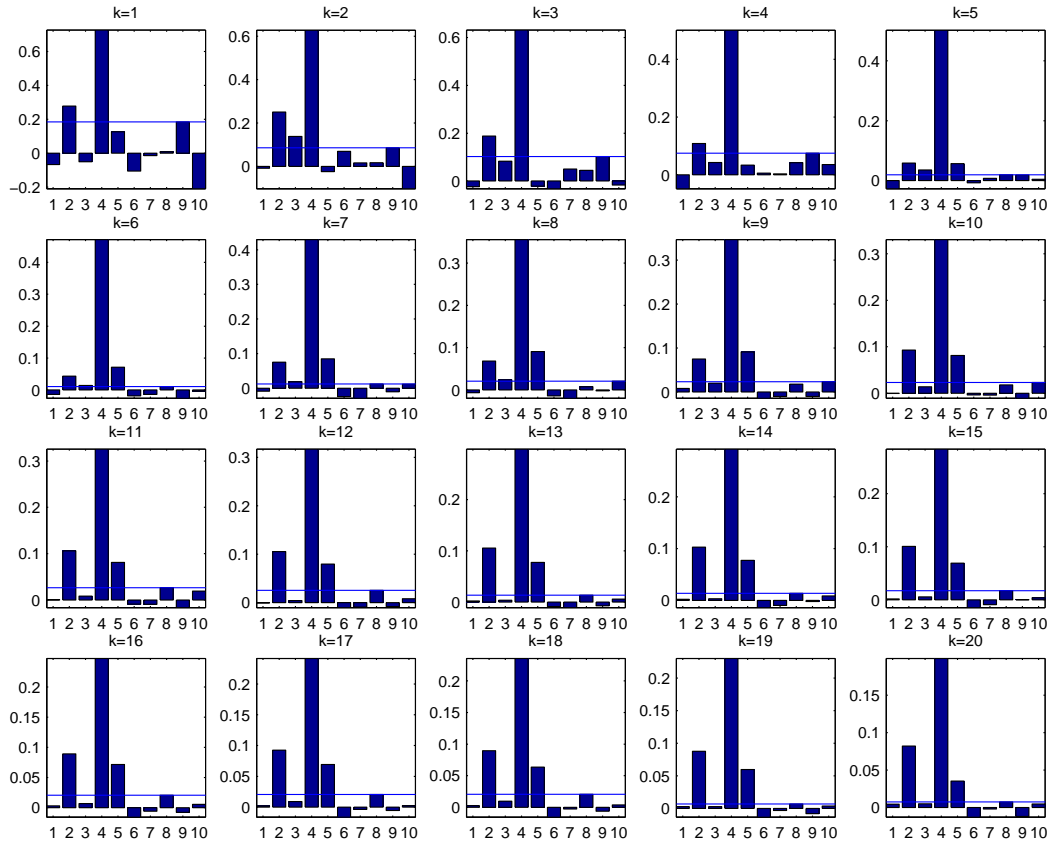}
  \caption{- Mutual information between the 10 variables of the synthetic example and the output, for several values of the estimator number of neighbors.  All relevant features have a higher mutual information than non-relevant ones only for well-chosen values.} 
  \label{fig:illustratechoiceofk}
\end{figure}

In practice, a bad choice of $k$ can modify the
ranking between variables and lead to false conclusions. As an illustration,
Figure \ref{fig:illustratechoiceofk} displays the mutual
information between each $X_i$ and $Y$, using the nearest neighbor-based
estimator for a single dataset generated from Equation \ref{eq:friedman}. The number
of neighbors used in the estimation of the mutual information is shown at the top of the graphs. Although only features $X_1$
to $X_5$ are informative, they do not always have a mutual information larger
than the other features. Furthermore, a significant, large, difference can be
observed between $X_1$ and $X_2$ while they have the same influence on the
output. 

This simple experiment shows that the number of neighbors must be
chosen correctly to avoid artefacts from the estimator, even in simple cases.

\subsection{Stopping criterion instability}

The stopping criterion of the forward search will determine how many features are selected.
When nested subsets of features are considered, as in the forward search, the mutual
information is theoretically a non-decreasing function of the subset size; it can only
increase or remain constant as more features are added. Maximizing the mutual
information therefore does not make sense: the whole feature subset will
always, in theory, have the largest mutual information with the value to
predict. 

\begin{figure}[htbp]
  \centering
\includegraphics[angle=270,width=1.0\textwidth]{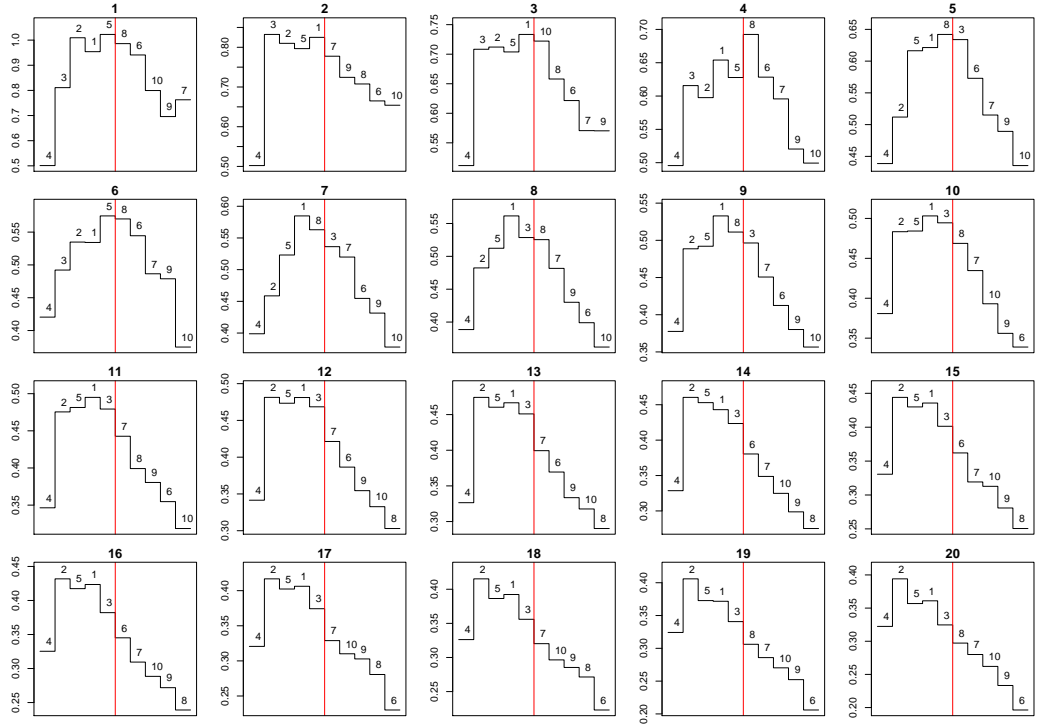}
  \caption{Result of the forward procedure on the artificial example with
    different values of the number of neighbors in the estimation of the mutual information. Only for well-chosen values of the number of neighbours the correct features ($X_1$ to $X_5$) are  selected.}
  \label{fig:Forward}
\end{figure}

In practice however, as illustrated in Figure \ref{fig:Forward}, the
evaluation of the mutual information tends to decrease when useless variables
are added, especially with an estimator based on the distances between
observations. It is therefore tempting to look for the maximum value of the
mutual information. But again, as shown in Figure \ref{fig:Forward}, this will
frequently lead to sub-optimal feature subsets. On this example, stopping the
forward procedure at the first peak selects a wrong number of features in
almost all cases. Moreover, searching for the global maximum does not improve
a lot the situation: the optimal set of features is selected only in three
cases (for $k$ equal to 1, 3, and 6). 

In fact, there is no particular reason for this strategy (maximization of the mutual information) to give optimal results when the mutual information is estimated via a distance-based method. Indeed, the forward procedure tends to add features in their relevance order. Moreover, when a feature is included in the current subset, it has the same individual importance in the distance calculations as each previously selected feature. As a consequence, the influence of the previous features, which might be more relevant than the last one, on the mutual information estimator tends to decrease. As shown in Figure \ref{fig:Forward}, there are many cases in which the first five features are the optimal ones and yet the mutual information is not maximal for the five feature set. In fact, the forward procedure only fails for $k$ equal to 5, 7 and 9, when it selects the irrelevant feature 8 before the relevant feature 3. While an optimal choice of $k$ should in theory prevent estimator problems to lead to bad estimations of the mutual information, and therefore rule out values 5, 7, and 9 for $k$, we cannot guarantee that the optimal feature subset will correspond to the highest value of the estimated mutual information. This is in fact more an intrinsic limitation of the chosen estimator than a problem of its tuning; it is in a sense the price we have to pay for an estimator that is able to handle higher-dimensional data.

There is thus a need for a sound stopping criterion of the forward search based on the mutual information, in addition to the optimal choice of $k$.

\section{Proposed methodology}\label{sec:proposedapproach}

\subsection{The number of neighbours}
\label{sec:neighborChoice}
In order to determine the optimal number of neighbors in the estimation of the
mutual information, the notion of optimality must be explicitly defined since
there is no obvious criterion that we could maximize or minimize. As already discussed, we do not
want to optimize the number of neighbors with respect to the performances of a
prediction model built with the variables chosen by the procedure, because
this would render the search procedure too time-consuming.

The goal is to discriminate between features that are relevant for the problem
and features that are useless. We therefore consider the optimal value of $k$
to be the value for which the separation between the relevant features and
an independent feature is maximum. Since the estimator of the mutual
information has some variance, it is important to take this variance into
account in measuring the separability. If we had access to the distribution of
the mutual information estimate over the data, we could calculate a separation
between $\MI{X;Y}$ and $\MI{U;Y}$ (considered as random variables) for an
important feature $X$ and an useless feature $U$.  

\begin{figure}[htbp]
  \centering
\includegraphics[angle=270,width=1.0\textwidth]{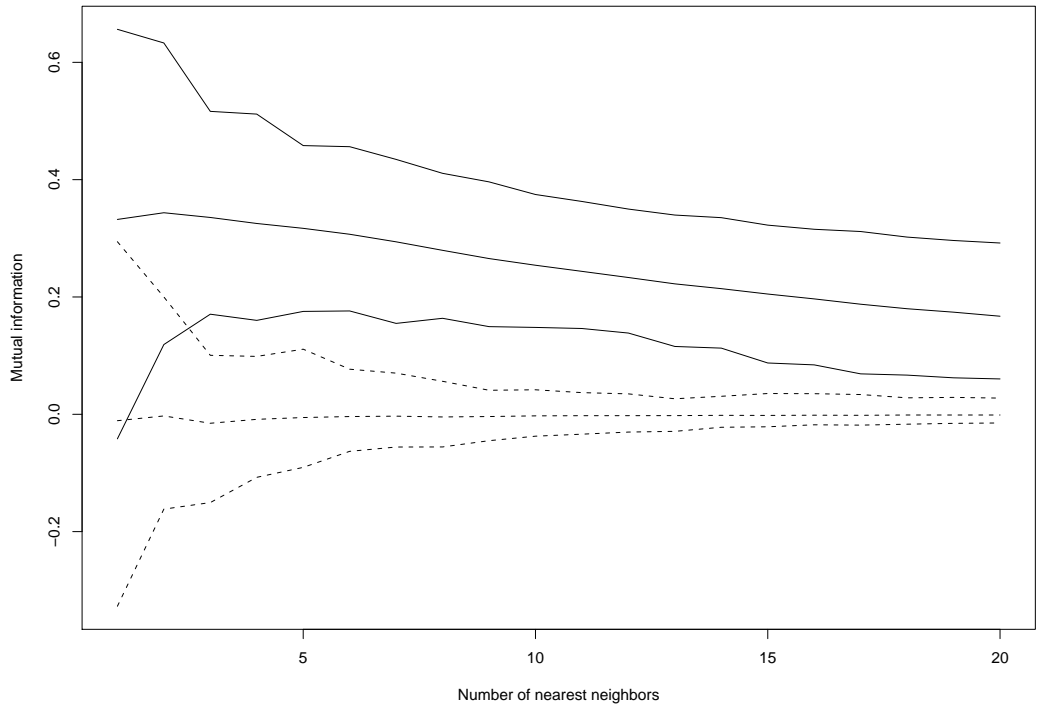}
\caption{Mutual information estimator distribution for datasets generated from
  Equation \ref{eq:friedman}. Solid lines correspond to variable $X_4$ and
  dashed lines to variable $X_{10}$. See text for details.}
  \label{fig:BiasVariance}
\end{figure}

To show the behavior of those variables on a simple example, 100 datasets are
randomly generated from Equation \ref{eq:friedman}. From those datasets, 100
realizations of the random variables $\MI{X_4;Y}$ and $\MI{X_{10};Y}$ are
produced, for different values of $k$. Figure \ref{fig:BiasVariance} represents
the means of $\MI{X_4;Y}$ and of $\MI{X_{10};Y}$ over the 100 datasets, as
well as the $0.01$ and $0.99$ percentiles of the same realizations. Those
values are reliable estimates of the theoretical values of the considered
quantities.

As expected, the variability of the estimator reduces with the number of
neighbors. However, the mutual information $\MI{X_4;Y}$ also decreases,
whereas there is a strong relationship between $X_4$ and $Y$. For a low number
of neighbors (1 and 2), the variability of the estimator is important enough
to blur the distinction between $X_4$ and $X_{10}$ in term of potential
predictive power: for some of the datasets, $\MI{X_{10};Y}$ is larger than $\MI{X_4;Y}$. When
$k$ increases, the estimator becomes stable enough to show that $Y$ depends
more on $X_4$ than on $X_{10}$ (for $k\geq 3$). However, after a first growing
phase, the separation between the distributions of $\MI{X_4;Y}$ and
$\MI{X_{10};Y}$ decreases with $k$: the reduction of the mean estimated value
of $\MI{X_4;Y}$ tends to negate the positive effect of the reduction of
variability. The lowest values of $\MI{X_4;Y}$ are getting closer and closer
to the highest values of $\MI{X_{10};Y}$. It seems therefore important to choose $k$ so as to ensure a good separation between relevant and irrelevant variables.

In practice however, the true distribution
of $\MI{X;Y}$ is unknown. We therefore rely on a combined K-fold/permutation
test to estimate the bias and the variance of the estimator for relevant
features and for independent ones. The idea is the following.  Consider $X_i$
a feature that is supposed to be relevant to predict $Y$. Two resampling
distributions are built for both $\MI{X_i;Y}$ and $\MI{X_i^\pi;Y}$ where
$X_i^\pi$ denotes a randomized $X_i$ that is made independent from $Y$ through
permutations. This is done by performing several estimations of (i) the mutual
information between $X_i$ and $Y$ and (ii) the mutual information between a
randomized version of $X_i$ and $Y$, using several non-overlapping subsets of
the original sample, in a K-fold resampling scheme. A good value
for $K$ is around 20 or 30. Less than 20 renders the estimation of mean and
variance hazardous, while the estimations with more than 30 are often very
close to those with K=30. The procedure results in two samples of estimates of
$\MI{X_i;Y}$ and $\MI{X_i^\pi;Y}$.

\begin{figure}[htbp]
\centering
\includegraphics[width=.65\textwidth]{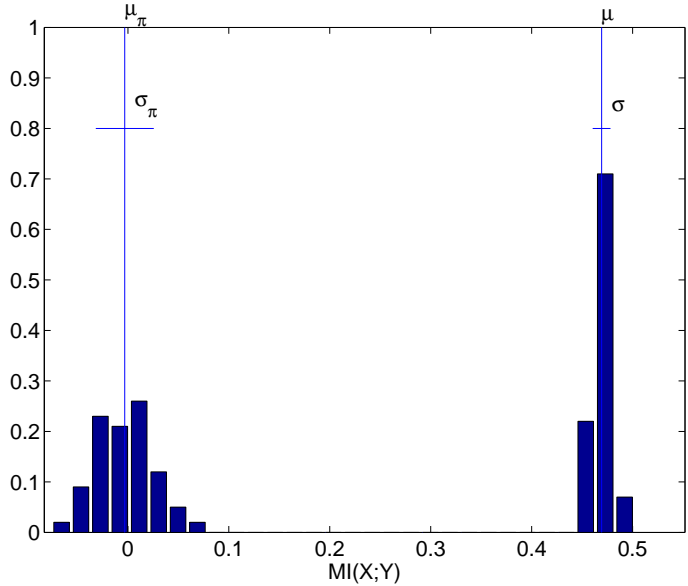}
  \caption{Distribution of mutual information for a relevant feature. On the left, the distribution of the mutual information of the features with permuted values, on the right, the distribution of the mutual information of the relevant feature; as given by the K-fold method. The value of $k$ is chosen so as to best separate those two distributions.} 
  \label{fig:illustratechoiceofkDistrib}
\end{figure}

The optimal value of $k$ is the one that best separates those two distributions, for instance according to a Student-like measure:
\begin{equation}
t_{i,k} = \frac{\mu-\mu_\pi}{\sqrt{\sigma^2 + \sigma^2_\pi}},
\end{equation}
where $\mu$ and $\sigma^2$ represent the mean and variance of the
cross-validated distribution of $\MI{X_i;Y}$, and $\mu_\pi$ and $\sigma^2_\pi$
are those of the cross-validated distribution of $\MI{X_i^\pi;Y}$ (illustrated
on Figure \ref{fig:illustratechoiceofkDistrib}).

The optimal $k$ for all features is chosen as the one corresponding to the
largest value of $t_{i,k}$ over all values of $k$ over all features. This way,
features that are useless do not participate in the choice of the optimal
value. Using useless features to choose the value that best separates the
resampling of the mutual information from the permuted sample would indeed
make no sense if they are independent from the output value.  It should
be noted that other solutions could be thought of, like, for instance, to
optimize the mean value of $t_{i,k}$ over features for which $t_{i,k}$ is
above a pre-specified significance threshold, but at the cost of an additional
parameter.

\subsection{The stopping criterion}\label{subsection:stoppingcriterion}

As choosing the maximum or the peak of the mutual information is nor sound neither efficient, 
a more promising approach consists in trying to avoid adding useless features
to the current subset by comparing the value of the mutual information with
the added feature to the one without that feature in a way that incorporates
the variability of the estimator.  

Let us
consider $S$, the subset of already selected features, and $X^*$, the best
candidate among all remaining features. We consider the distribution of
$MI\left( S \cup \{X^*\};Y \right)$ under the hypothesis that $X^*$ is
independent from $Y$ and $S$, that is all values of $MI\left( S \cup
  \{X^{*\pi}\};Y \right)$ where $X^{*\pi}$ is a random permutation of $X^*$.
If the P-value of $MI\left( S \cup \{X^*\} ;Y\right)$ is small and the
hypothesis is rejected, it means that $X^*$ brings sufficient new information
about $Y$ to be added to the feature subset.

Note that this way, the increment in mutual information between $MI\left( S
  \cup \{X^*\};Y \right)$ and $MI\left( S ;Y\right)$ is estimated without
comparing estimations of mutual information on subsets of different sizes. In
theory this should not be an issue; in practice however, it is important. Indeed, as we observed before, adding an
informative variable should, in theory, strictly increase the mutual information, but the
contrary is frequently observed (see for example Figure \ref{fig:Forward}.) 

\begin{figure}[htbp]
\centering
\includegraphics[width=.65\textwidth]{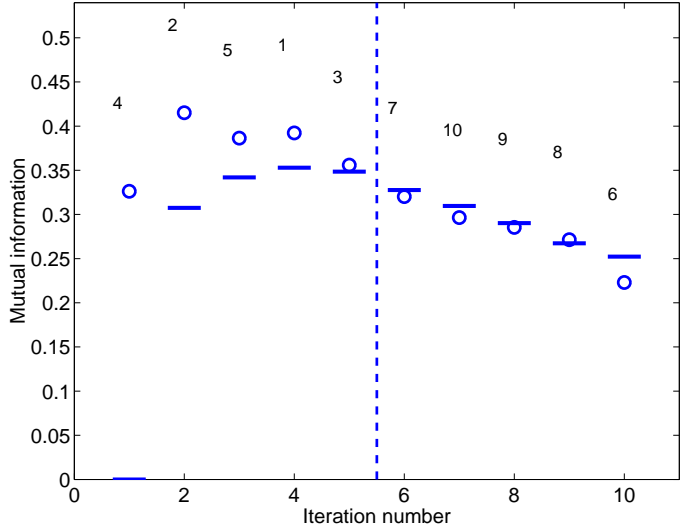}
  \caption{Mutual information in a forward feature subset
    search on the toy example. Thresholds (horizontal lines) are computed as
    the 95\% percentiles of the permutation distribution; the actual mutual information
    is represented with circles. The number of neighbors is $k=19$ (selected
    according to the criterion proposed in section \ref{sec:neighborChoice}). } 
  \label{fig:forwardtoy}
\end{figure}

Figure \ref{fig:forwardtoy} summarizes the results of the proposed stopping
criterion applied to the synthetic dataset introduced above. The procedure
selects the right features ($X_1$ to $X_5$) and finds that the sixth added
feature does not improve the mutual information significantly. As already shown on
Figure \ref{fig:Forward}, the mutual information decreases when the third
feature is added, which can wrongfully be taken as a clue that the procedure
should be halted. The permutation test is able to cope with the instabilities
of the estimator and to detect the relevance of the added feature even if it
makes the mutual information decrease. 

\subsection{Computational burden}
In most traditional resampling schemes, the overall computation time is simply multiplied by the number of resamplings performed. In this case however, a more detailed analysis is needed to grasp the overhead cost brought by the proposed method.

The number of mutual information estimations to perform at iteration $t$ in
the forward search, is equal to the number $d-t+1$ of features that are
candidate for entering the optimal feature subset plus the number $P$ of
permutations performed to evaluate the threshold of the stopping criterion.
The cost of each iteration, in terms of mutual information estimation, thus
amounts to $ d-t+1+P$. As the number of permutations is often limited to 20 or
30, the additional cost at each iteration needed to estimate the threshold is
small compared with the cost needed to find the feature that should be added
to the optimal feature subset. For instance, on a 100-dimensional dataset
(like the Delve census dataset presented in Section \ref{sec:delve}), 955
estimations of the mutual information are needed to find the optimal subset of
size 10 while 200 estimations, that is a bit more than 15\% were used to set
the threshold. Of course, when the number of original features is small,
permutations tend to represent a more important part of the total
computational burden.

The cost of the choice of the optimal number of neighbors is roughly equal to the
cost of the first step of the forward search multiplied by $K$, the number of
folds in the cross-validation scheme used in the proposed method. In practice,
$K$ is chosen between 20 and 30. If the expected number of optimal features
has the same order of magnitude, the total cost of the forward procedure will
also be of the same order of magnitude than the cost of the cross-validation,
which means that the overall cost is roughly doubled. However, this is much less
than if the number of neighbors was optimized using the performances of the
prediction model, as this would imply performing as many forward searches as
the number of values that are tested. 

The total cost of the automatic determination of the parameters, in the case of high-dimensional data, is thus a bit more than the double of the cost when the number of neighbors is chosen arbitrarily and the mutual information is maximized. This additional cost brings in better and more stable results, as shown in the next section.

\section{Experiments}\label{sec:experiments}

This section presents further experiments on the synthetic example and on three real-world datasets.

\subsection{A simulation study}
To further validate the interest of the proposed approach, the forward
procedure is applied to 100 datasets randomly generated from Equation
\ref{eq:friedman}. For each dataset, the optimal value of $k$ is selected
between 1 and 20, then the forward procedure is conducted. The feature set
that maximizes the mutual information and the best feature set according to
the stopping criterion presented in the previous section are retained for comparison. Results
are summarized in Tables \ref{table:featuresubsetsize:friedman},
\ref{table:informativefeaturesubsetsize:friedman} and
\ref{table:uselessfeaturesubsetsize:friedman}.

\begin{table}[htbp]
  \centering
  \begin{tabular}{lcccccc}\hline
Number of features & 1 & 2 & 3 & 4 & 5 &6 \\\hline
Maximal mutual information  & 7 &45 &33 &14 & 1 & 0\\
Stopping criterion & 0&1 &12 &52 &29 & 6\\\hline
  \end{tabular}
  \caption{Number of feature subsets of a given size obtained by both criteria}
\label{table:featuresubsetsize:friedman}
\end{table}

It appears clearly from Table \ref{table:featuresubsetsize:friedman} that
maximizing the mutual information does not provide good results: this leads to
the selection of an optimal set of features (5 variables) only in one case out
of one hundred. The stopping criterion defined in Section
\ref{subsection:stoppingcriterion} tends to select more features: in fact, the
feature sets obtained by this methods have strictly more features that the
ones selected by maximizing the mutual information in 84 \% of the cases (and
equal sizes in other situations).

\begin{table}[htbp]
  \centering
  \begin{tabular}{lcccccc}\hline
Number of informative features & 1 & 2 & 3 & 4 & 5 \\\hline
Maximal mutual information  & 7 &45 &33 &14 & 1 \\
Stopping criterion & 0&1 &16&66 &17 \\\hline
  \end{tabular}
  \caption{Number of feature subsets that contain the specified number of
    relevant features obtained by both criteria}
\label{table:informativefeaturesubsetsize:friedman}
\end{table}

Moreover, the additional features are generally informative ones, as
illustrated by Table \ref{table:informativefeaturesubsetsize:friedman}. The
positive aspect of maximizing the mutual information is that it leads, on
those experiments, only to the selection of relevant features. The stopping
criterion proposed in Section \ref{subsection:stoppingcriterion} selects
sometimes irrelevant features (see Table
\ref{table:uselessfeaturesubsetsize:friedman}), but it also selects always at
least as much relevant features as the former method. Moreover, in 79\% of the
experiments, it selects strictly more relevant features than the maximizing
strategy. In 5 \% of the experiments, the feature set selected by the
significance stopping criterion consists in the set that maximizes the mutual
information with an additional uninformative variable: this corresponds to the
error level expected as the forward procedure was controlled by using the 95\%
percentile of the permutation distribution.

\begin{table}[htbp]
  \centering
  \begin{tabular}{lccc}\hline
Number of uninformative features & 0 & 1 & 2 \\\hline
Stopping criterion & 75&22 &3 \\\hline
  \end{tabular}
  \caption{Number of feature subsets that contain the specified number of
    irrelevant features obtained by the stopping criterion of Section
    \ref{subsection:stoppingcriterion}}
\label{table:uselessfeaturesubsetsize:friedman}
\end{table}

This simulation study shows that while the proposed stopping criterion is not
perfect, it provides significant improvements over the standard practice of maximizing
the mutual information. Moreover, it does not lead to the selection of too
large feature sets that would reduce its practical benefit. 

The utility of the method is further illustrated below on a well-known dataset from the UCI Machine
Learning Repository (Housing), on a high-dimensional nitrogen spectra data and on a high-dimensional data set from the Delve repository.

\subsection{The HOUSING dataset}
 The goal with the Housing dataset is to predict the value of
houses (in k\$) described by 13 attributes representing demographic statistics
of the area around each house. The dataset contains 506 instances split into 338
learning examples and 169 test ones.

The optimal value (on the learning set) of $k$, searched between 1 and 20, is
found to be 18.

The forward search procedure described in the previous section is run with 50 permutations on the learning examples.  The threshold p-value is set to 0.05. When the mutual information is below the 95\% percentile of the permutation distribution, the procedure is halted.

\begin{figure}[htbp]
\centering
\includegraphics[width=.65\textwidth]{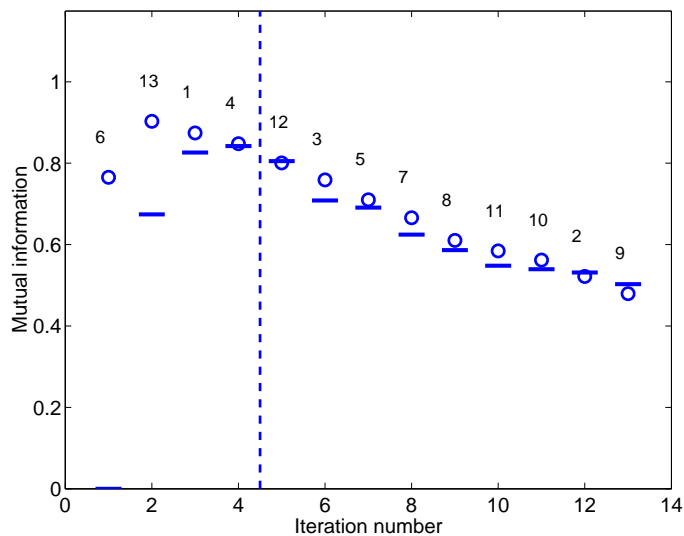}
  \caption{The evolution of the mutual information in a forward feature subset search on the Boston Housing dataset. Thresholds (horizontal lines) are computed as the 95\% percentile of the permutation distribution; the actual values of the  mutual information are represented with circles. The procedure stops after four features have been selected (dashed line).}
  \label{fig:forwardhousing}
\end{figure}

Figure \ref{fig:forwardhousing} displays the mutual information as a function
of the forward search iterations. The horizontal lines correspond to the
critical values (i.e. the 95\% percentile of the permutation distribution)
while the circles represent the mutual information between the selected
subset and the value to predict. Four features are selected ($X_6$, $X_{13}$,
$X_1$ and $X_4$). Interestingly, the procedure does not stop when the peak in
mutual information is observed.

A RBFN model was built using the selected features and optimized by 5-fold
cross-validation on the learning set, according to the method described in
\cite{Benoudjit:2003}. The root mean squared error (RMSE) on the test set is
9.48. By comparison, the RMSE on the test set with the all set of features is 18.97, while
the RMSE with the first two features, corresponding to the peak in mutual
information, is 19.39.

\subsection{The nitrogen dataset}
The nitrogen dataset originates from a software contest organized at the
International Diffuse Reflectance
Conference\footnote{\url{http://www.idrc-chambersburg.org/index.htm}} held in
1998 in Chambersburg, Pennsylvania, USA. It consists of scans and chemistry
gathered from fescue grass (\emph{Festuca elatior}). The data set contains 141
spectra discretized to 1050 different wavelengths, from 400 to 2 498 nm. The
goal is to determine the nitrogen content of the grass samples (ranging from
0.8 to 1.7 approximately). The data can be obtained from the Analytical
Spectroscopy Research Group of the University of
Kentucky\footnote{\url{http://kerouac.pharm.uky.edu/asrg/cnirs/shoot_out_1998/}}.

The dataset is split into a test set containing 36 spectra and a training set
with the remaining 105 spectra. We apply moreover a functional preprocessing,
as proposed in \cite{RossiEtAl05Neurocomputing}: this consists in replacing
each spectrum by its coordinates on a B-spline basis, which is itself
selected by minimizing a leave-one-out criterion (see
\cite{RossiEtAl05Neurocomputing} for details). The purpose of this functional
preprocessing is to reduce the huge number of original features (1050) to a
more reasonable number: the optimal B-spline basis consists indeed in 166
 B-splines of order four. 

\begin{figure}[htbp]
\centering
\includegraphics[width=.65\textwidth]{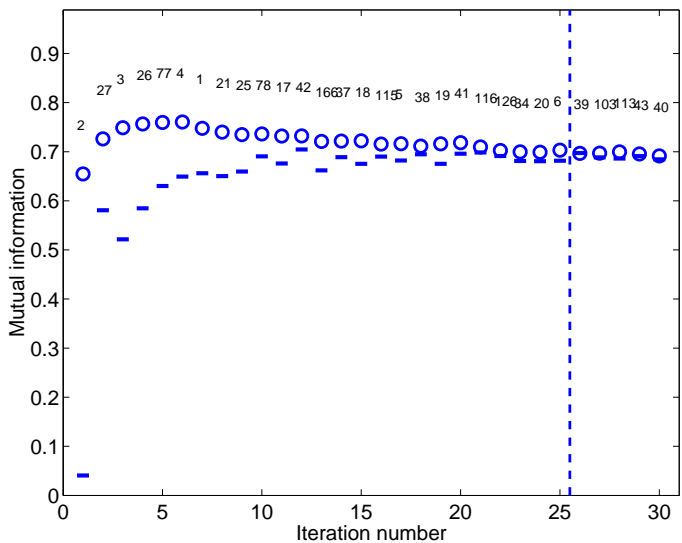}
  \caption{The evolution of the mutual information in a forward feature subset search on the nitrogen dataset. Thresholds (horizontal lines) are computed as the 95\% percentile of the permutation distribution; the actual values of the  mutual information are represented with circles. Twenty-five features are selected.}
  \label{fig:forwardnitro}
\end{figure}

Figure \ref{fig:forwardnitro} illustrates the behavior of the forward feature
selection with resampling on this dataset. The optimal number of neighbors is
12. It leads to the selection of 25 variables (among the 166 B-spline
coordinates). The RMSE on the test set, using a RBFN model built on those features,
is 0.6649. 

Maximizing the mutual information leads to a smaller feature set with 6
features. The RMSE on the test, using a RBFN model built on those features,
is 0.7753. As a reference, the RMSE on the test set when all features are used is
3.1197. 

\subsection{The Delve-Census dataest}\label{sec:delve}

The Delve Census dataset, available directly from the University of Toronto\footnote{\url{http://www.cs.toronto.edu/~delve/data/census-house/desc.html}}, comprises data collected during the 1990 US Census. Each of 22,784 the data elements concerns a small survey region and is described by 139 features measuring demographic information like the total person count in the region, the proportion of males, the percentage of people aged between 25 and 64, etc. The aim is to predict the median price of the houses in each survey region. This problem can be considered as a large scale version of the Housing dataset. 

For the sake of this analysis, we used only 104 of the 139 original
features. We indeed discarded the features that are too much correlated with
the value to predict like the average price, the first and third percentiles,
etc. In the dataset, 52 regions were found to have a median house price of
zero; they were considered to be erroneous  and removed from the analysis\footnote{The preprocessed data can be downloaded from the UCL Machine Learning Group website: \url{http://www.ucl.ac.be/mlg}}.

Of the 22,732 remaining observations, 14,540 are used for the test set. The 8192 remaining observations are split into 8 subsets and used for training. This corresponds to the classical splitting for this dataset; it allows one to study the variability of the feature selection procedure while retaining enough data both for learning and testing. For each observation subset, the optimal feature subset is determined using the proposed approach and a RBFN model is built using a 3-fold cross validation procedure. The RBFN model is then applied on the test set and the results are compared with those obtained using the peak in mutual information and using all features.

\begin{figure}[htbp]
\centering
\includegraphics[width=.65\textwidth]{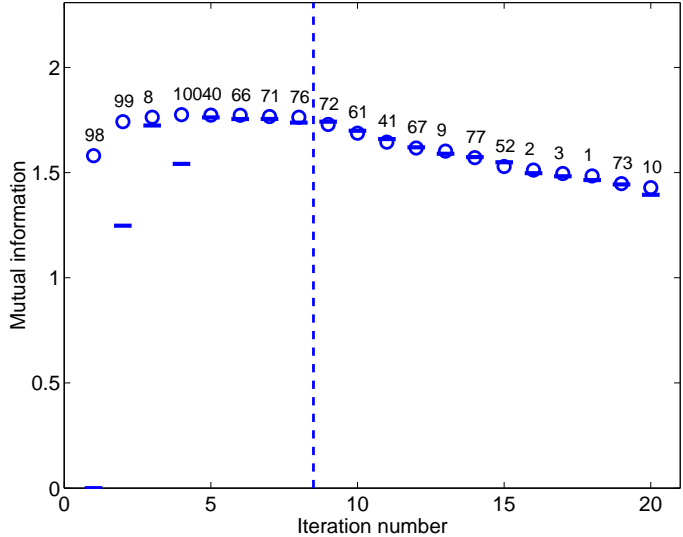}
  \caption{The evolution of the mutual information in a forward feature subset search on the first subset of the Delve dataset. Thresholds (horizontal lines) are computed as the 95\% percentile of the permutation distribution; the actual values of the mutual information are represented with circles. Eight features are selected.}
  \label{fig:forwarddelve}
\end{figure}

Figure \ref{fig:forwarddelve} displays the evolution of the mutual information and of the thresholds found by permutation over each iteration of the forward search procedure. Figure \ref{fig:forwarddelve} shows the results of the first of the eight learning sets. The number of selected features is eight, while the maximum of mutual information is observed for six features.

Table \ref{table:resultsdelve} shows the Root Mean Square Error of the model on the test set,  for each learning subset. The permutation approach always selects either 8 or 9 features, while stopping the forward procedure at the peak of mutual information gives from 2 to 6 features. Except for Subset number 2, the results obtained with the permutation are either equivalent either far better than those obtained with features selected by taking the peak of mutual information.

\begin{table}[htbp]
  \centering
  \begin{tabular}{p{1.5cm}|cp{2cm}|cp{2cm}|p{2cm}}\hline
           & \multicolumn{2}{l|}{Using permutations} & \multicolumn{2}{l|}{Using the peak} & All features\\
  Subset number & \# features  & test RMSE ($\times 10^4$) &\# features & test RMSE ($\times 10^4$) &test RMSE ($\times 10^4$)\\\hline
1&	8&	1.3286	&6	&1.3223	&	1.4304 \\
2&	9&	1.0748	&6	&0.9472	&	1.5393\\
3&	8&	1.2883	&3	&2.5643	&	1.4338\\
4&	8&	1.2214	&2	&2.3125	&	1.419\\
5&	9&	1.2575	&3	&1.1799	&	1.4628\\
6&	8&	0.9504	&5	&2.363	&	1.4146\\
7&	8&	1.1987	&2	&2.2381	&	2.1855\\
8&	9&	1.1929	&3	&1.19		& 1.5314\\\hline
  \end{tabular}
  \caption{Root mean square error on the test set obtained by the RBFN built on each learning subset. }
\label{table:resultsdelve}
\end{table}

\subsection{Discussion}
The three real-world examples illustrate the gain in prediction performances that can be obtained when using a well-chosen subset of features.  Simulations show the significant improvements obtained when using the proposed method for selecting the subset, rather than using as traditionally the peak of the mutual information, or the full set of features.

It appears therefore that the proposed strategy allows the automatic selection of
 good subsets of the original feature set. Moreover, it could easily be combined
with a simple wrapper approach that  compares the feature set that maximizes the mutual
information with the one obtained by the proposed method. This would further increase the
robustness of the feature selection process without leading to the enormous
computation time that would be required by a full wrapper forward selection
process.

\section{Conclusions}\label{sec:conclusions}

Combining the use of the mutual information and a forward procedure is a good
option for feature selection.  It is indeed faster than a wrapper approach
(that uses the prediction model itself for all evaluations) and still make
very few assumptions about the data as it is nonlinear and nonparametric.  The
major drawback of this approach is that the estimation of the mutual
information is often difficult in high-dimensional spaces, i.e. when several
features have already been selected.

Nearest neighbor-based mutual information estimators are one of the few sustainable
options for such estimation. However, two issues must be addressed. The first
one is the choice of the parameter of the estimator, namely the number of
neighbors. This number must be chosen carefully, especially with
high-dimensional subsets. The second one is the number of features to select,
or, equivalently, when to halt the forward procedure.

These two parameters of the approach could be optimized with respect to the performances of the prediction model, but this would require a large amount of computations. Rather, resampling methods can be used.

In this paper, the K-fold  and permutation resamplings are used in a combined way to
obtain an estimate of the variance of the estimator both in the case of
relevant features and of independent ones. The optimal number of neighbors
is then chosen so as to maximize the separation between the two cases.

Once the number of neighbors is chosen, the forward procedure may begin. It is
halted when the added feature does not significantly increase the mutual
information compared with the estimation of the mutual information if the same
feature was independent from the value to predict. This is done using the permutation
test. 

Combining the forward feature selection procedure, the mutual information to
estimate the relevance of the input subsets and resampling methods to estimate
the reliability of the estimation thus brings a feature selection methodology
that is faster than a wrapper approach and only requires the user to choose a
significance level; all other parameters are set in an automated way.

The method is illustrated on a synthetic dataset, as well as on three
real-world examples. The method is shown to perform
better than choosing the peak in mutual information. The test error of a
Radial Basis Function Network built with the features selected by the method
is always much lower than if the whole set of features is used and  significantly
lower than if  the features up to the peak in mutual information are used.

Although the procedure described here uses a forward feature selection, it
could be used as well with other incremental search methods like backward feature
elimination, or add-$r$ remove-$s$ methods that remove and/or add several
features at each step. Adaptive methods could be used also to detect when
performing more permutations is not necessary (for instance the variance in
the permuted data gets to a stable value). Furthermore, this paper focusses on mutual information
because it has been shown to be well adapted to forward feature selection, but
the methodology could be applied to quadratic mutual information
\cite{Principe} or to the Gamma test \cite{Jones} as well.

\section*{\small Acknowledgments}
D. Fran\c{c}ois is funded by a grant from the Belgian
F.R.I.A. Parts of this research result from the Belgian Program on Interuniversity
Attraction Poles, initiated by the Belgian Federal Science Policy Office.
The scientific responsibility rests with its authors.

\bibliographystyle{elsart-num}
\bibliography{PermutationTest}

\end{document}